\documentclass[letterpaper, 10 pt, conference]{ieeeconf}  

\usepackage{cite} 
\usepackage{graphicx}
\usepackage{booktabs}
\usepackage{nicematrix}
\usepackage{amsmath}
\usepackage{amssymb}
\usepackage{siunitx}
\usepackage{tikz} 
\newcommand*\circled[1]{\tikz[baseline=(char.base)]{
            \node[shape=circle,draw,inner sep=0.5pt] (char) {#1};}}

\usepackage{hyperref}
\hypersetup{
    colorlinks=false, 
    pdfborderstyle={/S/S/W 1}, 
    linkbordercolor={0 1 0},   
    citebordercolor={0 1 0},   
    urlbordercolor={0 1 0},    
}

\usepackage{xcolor}

\definecolor{expertBlack}{rgb}{0.0, 0.0, 0.0}        
\definecolor{expertYellow}{rgb}{1.0, 0.863, 0.196}   
\definecolor{expertMagenta}{rgb}{0.784, 0.392, 1.0}  
\definecolor{expertRed}{rgb}{0.941, 0.314, 0.314}    
\definecolor{expertBlue}{rgb}{0.0, 0.588, 1.0}         
\definecolor{expertGreen}{rgb}{0.392, 0.784, 0.0}     

\usepackage[font=normalsize]{caption}
            
\IEEEoverridecommandlockouts                              

\title{\LARGE \bf 
    \raisebox{-0.15\height}{\includegraphics[height=2em]{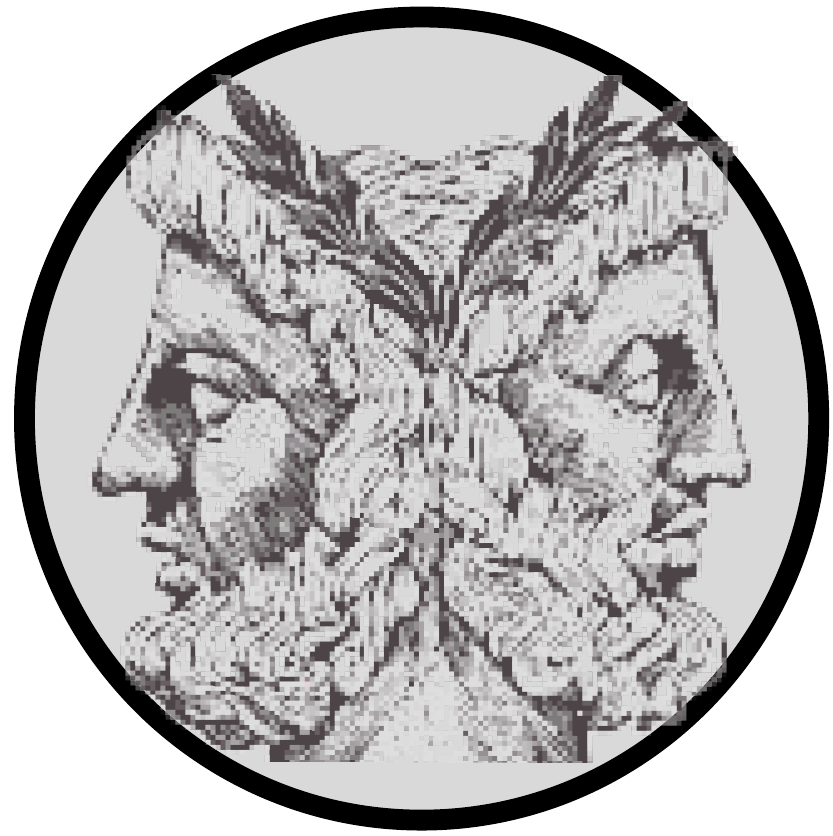}}\hspace{0.2em}GEMINUS: Dual-aware Global and Scene-Adaptive Mixture-of-Experts for End-to-End Autonomous Driving
}

\author{Chi Wan$^{1}$, Yixin Cui$^{1}$, Jiatong Du$^{1}$, Shuo Yang$^{1}$, Yulong Bai$^{1}$, Peng Yi$^{2}$, Nan Li$^{1}$, Yanjun Huang$^{1}$ 
\thanks{This work was supported by the National Natural Science Foundation of China, Joint Fund for Innovative Enterprise Development (U23B2061).(Corresponding author: Yanjun Huang.)}
\thanks{$^{1}$School of Automotive Studies, Tongji University, Shanghai, China.
        {\tt\small \{2532900, 2411448, 2210197, 2111550, 2210197, li\_nan, yanjun\_huang\}@tongji.edu.cn}}
\thanks{$^{2}$School of Electronics and Information Engineering, Control Science and Engineering Department, Tongji University, Shanghai 200070, China.
        {\tt\small yipeng@tongji.edu.cn}}%
}

\begin{document}

\maketitle
\thispagestyle{empty}
\pagestyle{empty}


\begin{abstract}

End-to-end autonomous driving requires adaptive and robust handling of complex and diverse traffic environments. However, prevalent single-mode planning methods attempt to learn an overall policy while struggling to acquire diversified driving skills to handle diverse scenarios. Therefore, this paper proposes GEMINUS, a Mixture-of-Experts end-to-end autonomous driving framework featuring a Global Expert and a Scene-Adaptive Experts Group, equipped with a Dual-aware Router. Specifically, the Global Expert is trained on the overall dataset, possessing robust performance. The Scene-Adaptive Experts are trained on corresponding scene subsets, achieving adaptive performance. The Dual-aware Router simultaneously considers scenario-level features and routing uncertainty to dynamically activate expert modules. Through the effective coupling of the Global Expert and the Scene-Adaptive Experts Group via the Dual-aware Router, GEMINUS achieves both adaptability and robustness across diverse scenarios. GEMINUS outperforms existing methods in the Bench2Drive closed-loop benchmark and achieves state-of-the-art performance in Driving Score and Success Rate, even with only monocular vision input. The code is available at \url{https://github.com/newbrains1/GEMINUS}.
\end{abstract}

\begin{figure*}[t]
\centering
\includegraphics[width=7.0in]{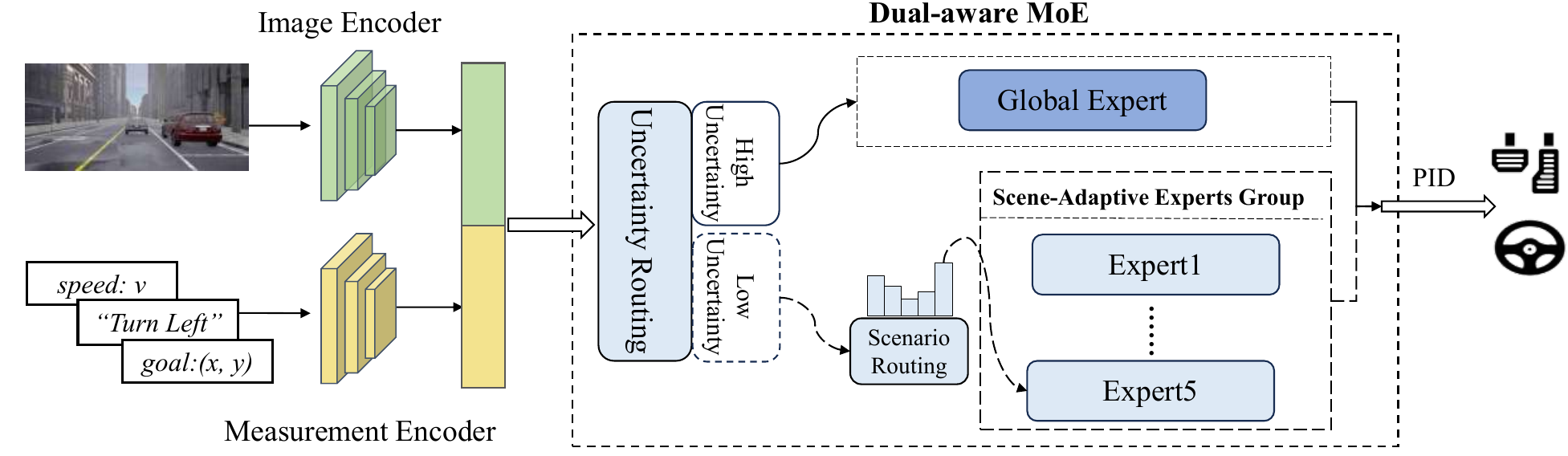}
\caption{\textbf{The overall architecture of GEMINUS.} GEMINUS integrates a Global Expert and a Scene-Adaptive Experts Group. During the training stage, the Global Expert is trained on the overall dataset and each scene-adaptive expert is trained on respective scene subset guided by scenario-aware routing. In the inference stage, as shown above, features extracted from upstream encoders are processed by the Dual-aware Router. When scenario uncertainty is below a threshold, the scenario-aware routing activates the highest-scoring adaptive expert; conversely, for high uncertainty, the uncertainty-aware routing activates the Global Expert. This design ensures both adaptive and robust performance across diverse scenarios.}
\label{GEMINUS} 
\end{figure*}

\begin{figure}[t]
\centering
\includegraphics[width=1\linewidth]{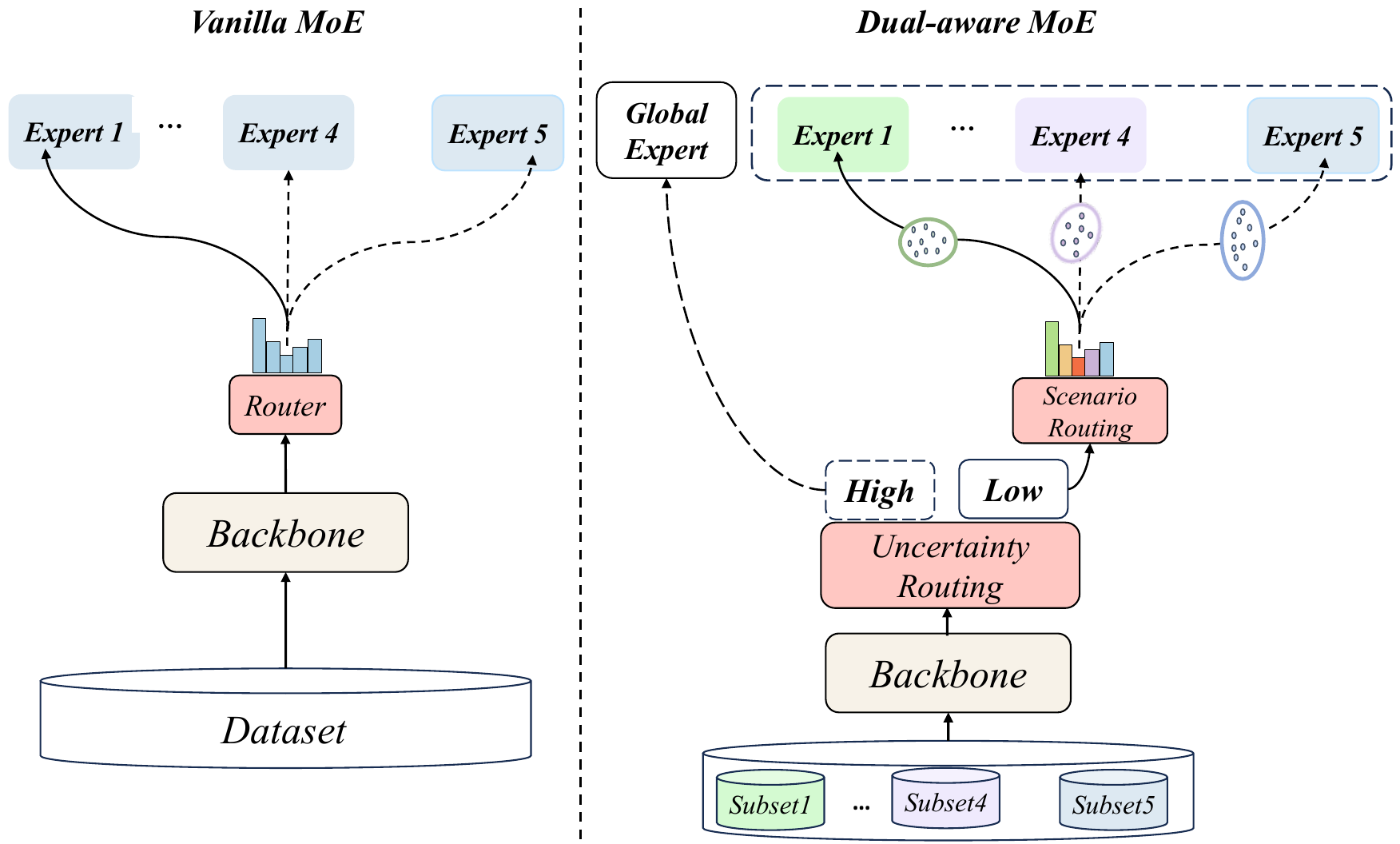}
\caption{\textbf{Dual-aware MoE vs Vanilla MoE.} Unlike a vanilla MoE which typically balances expert load, the Dual-aware MoE employs two distinct routing strategies: scenario-aware routing and uncertainty-aware routing. The scenario-aware routing leverages subset IDs during training to guide specific inputs to corresponding experts, enabling the learned router to dynamically activate the most appropriate expert from scene features alone at inference. Furthermore, to ensure robustness in scenarios with ambiguous features, we further introduce an uncertainty-aware routing and a global expert.}
\label{DualMoE}
\end{figure}

\section{INTRODUCTION}

In recent years, a prominent research direction in autonomous driving is the development of planning-oriented end-to-end models\cite{end2endreview}. In contrast to modular autonomous driving, which consists of a pipeline of perception, prediction and planning
\cite{bevformer,zhang2025co,pathplanning}, end-to-end methods directly map raw sensor inputs to planned trajectories \cite{transfuser,uniad,vad,drivetransformer}, control signals \cite{conditionalimitationlearning,driveadapter}, or a fused output derived from trajectory and control branches \cite{tcp,multiadaptive}. These approaches provide a holistic model for driving, enabling unified optimization towards a global objective, significantly reducing manual engineering efforts, and allowing for the direct use of rich sensor information.

Despite their notable benefits, a persistent limitation of current end-to-end autonomous driving models stems from their global imitation learning on overall training datasets. This approach, typically employing single-mode planning with L2 loss, inherently models the complex output space as a single Gaussian distribution, leading to a tendency toward mode averaging\cite{drivetransformer,motionforecasting}. Consequently, their performance is compromised, as the generated output represents an averaged behavior across diverse scenarios rather than the optimal policy for the current specific scenario. This ultimately restricts the acquisition of diversified driving skills to handle diverse scenarios. Prior approaches employed command-based conditional imitation learning to mitigate mode averaging\cite{conditionalimitationlearning,CILRS}. However, this approach faced an inherent limitation: solely relying on driving commands is insufficient to distinguish complex scenarios (e.g., an overtaking scenario simultaneously includes turn left, go straight, and turn right commands). Such rigid classification fails to comprehensively consider rich scene information, thus hindering the capture of the diversity of driving skills.

Inspired by the success of Mixture-of-Experts (MoE) architectures in large language models (LLMs) to handle complex data distribution\cite{moesurvey}, MoE architectures present significant potential for addressing challenges in autonomous driving. By offering a fine-grained scenario adaptation and specialized behavior generation, MoE could mitigate the mode averaging problem and enhance model adaptability in diverse driving scenarios. However, directly transferring generic MoE architectures, designed primarily for static textual data, to autonomous driving reveals an inherent unsuitability. Specifically, they struggle with effective expert specialization due to lack of explicit scenario division and fail to adequately consider the robustness requirements of autonomous driving.

Therefore, this paper proposes GEMINUS: dual-aware \underline{\textbf{G}}lobal and sc\underline{\textbf{E}}ne-adaptive \underline{\textbf{MI}}xture of experts for end-to-end auto\underline{\textbf{N}}omo\underline{\textbf{US}} driving. Our contributions can be summarized as follows:

\begin{itemize}

\item We present GEMINUS, a novel MoE framework for end-to-end autonomous driving. This framework adopts a cooperative architecture that partitions the driving policy between a robust Global Expert and a Scene-Adaptive Experts Group to simultaneously achieve adaptive performance in feature-distinct scenarios and robust performance in feature-ambiguous scenarios.
\item We devise a dual-aware Router as the core mechanism of GEMINUS. It is engineered with two complementary awareness capabilities: scenario awareness for precise expert dispatching in distinct situations and uncertainty awareness for ensuring robust fallback in ambiguous conditions.
\item We conducted in-depth analyses of GEMINUS's intrinsic routing mechanism
on the Bench2Drive benchmark, examining the impact of the uncertainty threshold on driving performance, as well as the router's accuracy and expert utilization.

\end{itemize}

\section{RELATED WORK}

\subsection{End-to-End Autonomous Driving}

End-to-end autonomous driving is an emerging paradigm that aims to directly map raw sensor inputs to vehicle control actions, thereby simplifying the system architecture and mitigating error propagation. Driving policies are predominantly learned via Imitation Learning (IL) to mimic expert demonstrations \cite{transfuser,uniad,vad,drivetransformer,conditionalimitationlearning,driveadapter,tcp,multiadaptive} or Reinforcement Learning (RL) for dynamic policy optimization \cite{learningdriveinaday,roach,think2drive}.

In 2022, Trajectory-guided Control Prediction (TCP) \cite{tcp} was proposed as a simple and robust monocular vision baseline. This method innovatively integrates trajectory planning and direct control into a unified pipeline for joint learning and predictive fusion. Furthermore, techniques such as TransFuser \cite{transfuser} adopt the Transformer to fuse information from cameras and LiDAR. Beyond fusion, recent innovations also include vectorized scene representations, exemplified by VAD \cite{vad}, which models driving scenes as fully vectorized elements to improve planning efficiency and robustness. Diffusion models have also emerged as a powerful tool, with DiffusionDrive \cite{diffusiondrive} utilizing truncated diffusion policies to model multi-modal action distributions while simultaneously achieving real-time control.

\subsection{Mixture-of-Experts in Autonomous Driving}

The MoE architecture has emerged as a significant method for scaling LLMs and enhancing task specialization. In LLMs, sparse MoE designs boost model capacity and processing efficiency through conditional computation \cite{moesurvey}. The strength of MoE lies in its ability to leverage individual experts' strengths across varying data subsets or tasks, thereby improving overall model performance. 

Despite promising results in LLMs, MoE's application in end-to-end autonomous driving remains underexplored. Some existing studies have explored MoE architectures in autonomous driving for tasks such as rare scenario perception \cite{raremoe}, long-tailed trajectory prediction\cite{amend}, domain adaptation in different weathers \cite{domainmoe}, safe trajectory prediction and planning \cite{safemoe}, continual adaptation to failure cases\cite{cui2025continual} and facilitating generalization of planner \cite{generalizingmoe}. However, these existing approaches have not focused on leveraging MoE to enhance adaptive and robust performance in diverse scenarios.

\begin{figure}[t]
\centering
\includegraphics[width=1\linewidth]{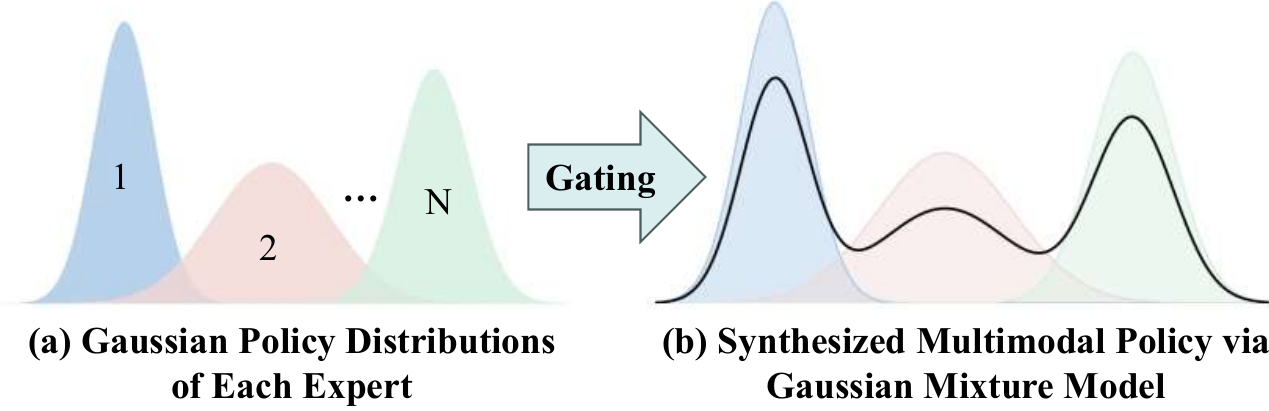}
\caption{\textbf{A Conceptual Illustration of Representing Multimodal Policy Distribution with Mixture-of-Experts.} }
\label{policy_mix}
\end{figure}

\section{METHODOLOGY}

Fig.~\ref{GEMINUS} illustrates the overall architecture of our proposed GEMINUS framework. Our approach begins with the construction of a foundational end-to-end model, inspired by the design philosophies of TCP \cite{tcp}. This model functions as a conventional single-expert baseline upon which we integrate our novel Dual-aware MoE architecture to form the complete GEMINUS framework.

\subsection{Preliminaries}
{\textbf{End-to-End Autonomous Driving.} The objective of end-to-end autonomous driving is to directly map raw sensor inputs to corresponding trajectories or control actions. In this paper, the raw sensor input $x$ encompasses: a front-facing camera image $i$, the ego-vehicle speed $v$, a high-level navigation command $c$, and a goal point $(x_g, y_g)$. The raw sensor inputs are processed by the end-to-end model. Encoders first process these inputs to generate intermediate features. These features are then fed into a trajectory planner. The trajectory planner generates a planned trajectory that comprises waypoints in $K$ steps. This planned trajectory is then fed into a Proportional-Integral-Derivative (PID) controller. The controller subsequently produces the final longitudinal control signals: throttle $\in[0,1]$, brake $\in[0,1]$, and the lateral control signal: steer $\in[-1,1]$.

\textbf{Mixture-of-Experts.} The MoE architecture offers a principled approach to address the complexities of multimodal data distributions employing a "divide and conquer" strategy \cite{moesurvey}. Introducing the MoE Framework to end-to-end trajectory planning, the overall policy distribution $p_{\theta}(Y \mid X)$ is typically represented as a probabilistic mixture of policy distributions of components $K$, each parameterized by an expert $m_{\theta}(Y \mid Z = k, X)$ and weighted by a gating network $q_{\theta}(Z = k \mid X)$, formalized as:

\begin{equation}
p_{\theta}(Y \mid X) = \sum_{k=1}^{K} q_{\theta}(Z = k \mid X) \cdot m_{\theta}(Y \mid Z = k, X)
\end{equation}

As conceptually illustrated in Fig.~\ref{policy_mix}, this probabilistic formulation allows a model to tackle the mode averaging problem by representing the multi-modal distribution of driving policies as a mixture of multiple expert policy distributions\cite{eda,online}. A common and effective simplification in deep learning implementations is to use a Hard Assignment approach, where a router selects a single, most suitable expert for a given input. This hard assignment strategy is highly effective as it avoids averaging the outputs of multiple experts, thus further mitigating mode averaging. Building upon these theoretical underpinnings, this paper proposes GEMINUS, a distinctive MoE framework that is specifically tailored for diverse and complex autonomous driving scenarios. At its core, a Dual-aware Router possesses scenario and uncertainty awareness to dynamically activate experts from a Global Expert and a Scene-Adaptive Experts Group. During inference, the Dual-aware Router processes intermediate features $x$ extracted by the encoders. It determines the final output $y$ based on the uncertainty measure $U(x)$ and the scores of scene experts $S_{E}(x)$, formalized as:

\begin{equation}
y = 
\begin{cases}
f_{\text{global}}(x), & \text{if } U(x) \geq \tau \\
f_{\operatorname*{argmax}\limits_{i \in \mathcal{S}} \; S_{E}(x)}(x), & \text{if } U(x) < \tau
\end{cases}
\end{equation}

Here $x$ is the input feature. $\tau$ denotes a predefined uncertainty threshold. $\mathcal{S}$ is the set of all the scene-adaptive experts. 
When $U(x)$ is low ($U(x)<\tau$), the expert with the highest routing score $S_{E_i}(x)$ is selected. This achieves precise and scenario-specific planning in feature-distinct scenarios. 
In contrast, when the uncertainty of the scenario $U(x)$ is high ($U(x)\geq\tau$), the model selects the Global Expert $f_{\text{global}}(x)$. This ensures robust performance in feature-ambiguous scenarios. This design allows GEMINUS to effectively avoid the mode averaging problem, thus achieving adaptive and robust performance in diverse scenarios.

\subsection{Single-Expert Baseline}

\textbf{Feature Encoders.} The image encoder is a ResNet34 architecture pretrained on ImageNet \cite{imagenet}, which processes the view of the front-facing camera. Concurrently, a measurement encoder processes vehicle state information, including its speed $v$, the navigation command $c$, and the goal point $(x_g, y_g)$. The feature vectors from both encoders are then concatenated into a combined feature vector $F$ for the planner.

\textbf{Trajectory Planner.} The trajectory planner first processes the combined feature $F$ through linear layers for downsampling. The resulting vector is then fed into a GRU-based model \cite{gru} that autoregressively generates a sequence of waypoints for the next $K=4$ steps. Finally, these waypoints are used by downstream controllers to produce throttle, brake, and steer commands.

\subsection{Scenario-aware Routing Mechanism}

Vanilla MoE aims to balance expert usage across GPUs to utilize maximum benefit from input features. However, this leads to inefficient knowledge sharing among experts when dealing with heterogeneous input distributions. For example, driving policies for a Merging scenario differ significantly from an Emergency Brake scenario. To address this inefficiency and foster specialized knowledge, a scenario-aware routing mechanism is introduced. This mechanism draws inspiration from dataset-aware routing in\cite{damex}.

Inspired by the scenario classification in Bench2Drive \cite{bench2drive}, five autonomous driving scenario categories are classified: Merging, Overtaking, Emergency Brake, Give Way, and Traffic Sign. During the training phase, the scenario-aware router is explicitly trained to route input feature vectors $F$ based on their corresponding scenario category.

Let $\mathcal{S}=\{s_m\}_{m=1}^{|\mathcal{S}|}$ be a set of predefined scenario categories. An input feature vector $x$ belongs to a scenario $s_m$ ($x\in s_m$). We define a mapping function $h:S\rightarrow E$. This function assigns each scenario category $s_m$ to a specific scene-adaptive expert $e_i\in E$. Here, $E$ denotes the group of scene-adaptive experts. To enforce this routing strategy, a router loss is designed as $L_{scenario}$. This loss is formulated as a cross-entropy loss. It is computed between the router's predicted expert selection probabilities $p_i(x)$ (representing the probability of selecting expert $e_i$) and the target expert label $h(s_m)$ corresponding to input $x$ from scenario $ s_m$:

\begin{equation}
\mathcal{L}_{\text{scenario}} = - \sum_{i=1}^{|E|} \mathbf{1}(h(s_m) = i) \cdot \log p_i(x) \label{scenario-aware}
\end{equation}

$\mathbf{1}(\cdot)$ is the indicator function. This loss ensures that all inputs originating from a specific scenario category are primarily dispatched to their designated scene-adaptive expert. By selectively routing inputs based on their scenario feature, this mechanism promotes efficient knowledge specialization within each expert. This enables the model to learn adaptive driving policies.

\subsection{Uncertainty-aware Routing Mechanism}

While scene-adaptive experts excel in feature-distinct scenarios, relying solely on them can be problematic in feature-ambiguous scenarios. This compromises robustness, especially in safety-critical applications like autonomous driving. To mitigate this risk and ensure reliable performance in diverse scenarios, an uncertainty-aware routing mechanism is introduced.

The router's predicted probabilities for the N experts, $p_i(x)$, as defined in the  preceding section, form a distribution $P(x)=[p_1(x), p_2(x), ..., p_N(x)]$.
Subsequently, the Information Entropy\cite{shannonentropy} of this distribution is calculated to reflect the uncertainty of the router's decision:

\begin{equation}
    H(P(x)) = - \sum_{i=1}^{N} p_i(x) \log(p_i(x))
\end{equation}

To normalize this entropy to the $[0,1]$ range, it is divided by the theoretical maximum entropy. This maximum occurs when probabilities are uniformly distributed across all experts (i.e., $p_i = 1/N$ for all $i$), and its value is $log(N)$. Thus, the Normalized Information Entropy $U(x)$ is defined as:

\begin{equation}
    U(x) = \frac{H(P(x))}{\log(N)} = \frac{-\sum_{i=1}^{N} p_i \log(p_i)}{\log(N)}
\end{equation}

This Normalized Information Entropy $U(x)$ serves as the measure of scenario uncertainty. A value close to 0 indicates high certainty, meaning the scenario is distinct and the router is confident. Conversely, a value close to 1 indicates high uncertainty, meaning that the scenario is ambiguous and the router is undecided.

\subsection{Loss Design}

GEMINUS is trained using a comprehensive loss function that combines multiple objectives. 

\textbf{Global Expert Loss.} The Global Expert aims to provide a robust and generalized driving policy. Its loss, $\mathcal{L}_{\text{Global}}$, is formulated as a weighted sum of three distinct components:
\begin{equation}
    \mathcal{L}_{\text{Global}} = \lambda_{\text{traj}} \, \mathcal{L}_{\text{traj\_global}} + \lambda_{F} \, \mathcal{L}_{F\_global} + \lambda_{V} \, \mathcal{L}_{V\_global}
\end{equation}
where $\mathcal{L}_{\text{traj\_global}}$ is a trajectory imitation loss that minimizes the L2 distance between predicted and ground-truth waypoints. The other two terms, a feature alignment loss $\mathcal{L}_{F\_global}$ and a value alignment loss  $\mathcal{L}_{V\_global}$, serve to distill knowledge from the Think2Drive expert \cite{think2drive} by minimizing the L2 distance to the teacher's intermediate features and predicted value. The $\lambda$ coefficients are tunable loss weights.

{\textbf{Scene-Adaptive Experts Group Loss.} The Scene-Adaptive Experts Group comprises $N$ distinct experts. Each expert is trained to master the policies for specific scenarios. The loss for this group, $L_{Adaptive}$, is calculated as a weighted sum of the losses of individual experts. For a given sample $x$, only the adaptive expert to which it is routed contributes to the loss. This calculation is similar to the components of the Global Expert Loss. It is formalized as:

\begin{equation}
\begin{split}
\mathcal{L}_{\text{Adaptive}} &= \mathbf{1}(x \rightarrow e_i) \cdot \Big( \lambda_{\text{traj}} \, \mathcal{L}_{\text{traj},i}(x) \\
&\quad + \lambda_F \, \mathcal{L}_{F,i}(x) + \lambda_V \, \mathcal{L}_{V,i}(x) \Big)
\end{split}
\end{equation}

For integers $i \in \{1, \dots, N\}$, $\mathcal{L}_{traj,i}(x)$, $\mathcal{L}_{F,i}(x)$, and $\mathcal{L}_{V,i}(x)$ are the trajectory imitation loss, feature alignment loss, and value prediction loss for expert $e_i$ on sample $x$. $\mathbf{1}(\cdot)$ is the indicator function that ensures that only the loss of the activated expert contributes to that specific sample.

\textbf{Total Loss.} The total loss function for training the GEMINUS model is the weighted sum of the expert losses, a router loss and an auxiliary speed prediction loss:
\begin{equation}
\begin{split}
\mathcal{L}_{\text{total}} &= \lambda_{\text{Global}} \, \mathcal{L}_{\text{Global}} + \lambda_{\text{Adaptive}} \, \mathcal{L}_{\text{Adaptive}} \\
&\quad + \lambda_{\text{scenario}} \, \mathcal{L}_{\text{scenario}} + \lambda_{\text{speed}} \, \mathcal{L}_{\text{speed}}
\end{split}
\end{equation}

Here, $\mathcal{L}_{\text{scenario}}$ is the router's cross-entropy loss used to train accurate expert selection (as described in Equation (\ref{scenario-aware}) ). $\mathcal{L}_{\text{speed}}$ is an L1 loss from a dedicated head that predicts the speed of the ego vehicle to improve its state estimation. The coefficients $\lambda$ are empirically determined weights that balance the contribution of each loss term.

\begin{table*}[t] 
\centering

\caption{Closed-Loop and Open-Loop Performance on Bench2Drive Benchmark.}
\label{tab1:Close and Open loop results}

\setlength{\tabcolsep}{6pt} 

\begin{tabular}{l|l|l|S[table-format=1.2]|S[table-format=2.2]|S[table-format=2.2]} 
\toprule
\textbf{Method} & \textbf{Venue} & \textbf{Input} & \multicolumn{1}{c|}{\textbf{Open-loop Metric}} & \multicolumn{2}{c}{\textbf{Closed-loop Metric}} \\ 
\cmidrule(lr){4-4} \cmidrule(lr){5-6} 
\textbf{} & \textbf{} & \textbf{} & {Avg. L2 (m)$\downarrow$} & {\textbf{Driving Score $\uparrow$}} & {Success Rate(\%) $\uparrow$} \\ 
\midrule
TCP* \cite{tcp} & NeurIPS 2022 & Ego State + Front Camera & 1.70 & 40.70 & 15.00 \\
TCP-ctrl* \cite{tcp} & NeurIPS 2022 & Ego State + Front Camera & {--} & 30.47 & 7.27 \\
TCP-traj* \cite{tcp} & NeurIPS 2022 & Ego State + Front Camera & 1.70 & 59.90 & 30.00 \\
UniAd-Base \cite{uniad} & CVPR 2023 & Ego State + 6 Cameras & 0.73 & 45.81 & 16.36 \\
ThinkTwice* \cite{thinktwice} & CVPR 2023 & Ego State + 6 Cameras & 0.95 & 62.44 & 31.23 \\
VAD \cite{vad} & ICCV 2023 & Ego State + 6 Cameras & 0.91 & 42.35 & 15.00 \\
DriveAdapter* \cite{driveadapter} & ICCV 2023 & Ego State + 6 Cameras & 1.01 & 64.22 & 33.08 \\
GenAD \cite{genad} & ECCV 2024 & Ego State + 6 Cameras & {--} & 44.81 & 15.90 \\
DriveTrans \cite{drivetransformer} & ICLR 2025 & Ego State + 6 Cameras & \textbf{0.62} & 63.46 & 35.01 \\
MomAD \cite{momenad} & CVPR 2025 & Ego State + 6 Cameras & 0.82 & 47.91 & 18.11 \\
SparseDrive \cite{sparsedrive} & ICRA 2025 & Ego State + 6 Cameras & 0.83 & 42.12 & 15.00 \\
TTOG \cite{TTOG}& ArXiv 2025 & Ego State + 6 Cameras & 0.74 & 45.23 & 16.36 \\
\midrule
GEMINUS* & {--} & Ego State + Front Camera & 1.60 & \textbf{65.39} & \textbf{37.73} \\
\bottomrule
\end{tabular}

\vspace{1ex} 
\begin{minipage}{\textwidth} 
\centering
\small
* denotes expert feature distillation. \\
Avg.L2 is averaged over the predictions in 2 seconds under 2Hz. 

\end{minipage}

\end{table*}

\begin{table*}[t]
\centering
\caption{MultiAbility Results on Bench2Drive Benchmark.}
\label{tab2 Multiability Results}

\setlength{\tabcolsep}{6pt}

\begin{tabular}{l|l|l|S|S|S|S|S|S}
\toprule
\textbf{Method} & \textbf{Venue} & \textbf{Input} & \multicolumn{6}{c}{\textbf{Ability (\%)$\uparrow$}} \\
\cmidrule(lr){4-9}
\textbf{} & \textbf{} & \textbf{} & {Merging} & {Overtaking} & {Em-Brake} & {Give Way} & {Traffic Sign} & {\textbf{Mean}} \\
\midrule
TCP* \cite{tcp} & NeurIPS 2022 & Ego State + Front Camera & 16.18 & 20.00 & 20.00 & 10.00 & 6.99 & 14.63 \\
TCP-ctrl* \cite{tcp} & NeurIPS 2022 & Ego State + Front Camera & 10.29 & 4.44 & 10.00 & 10.00 & 6.45 & 8.23 \\
TCP-traj* \cite{tcp} & NeurIPS 2022 & Ego State + Front Camera & 8.89 & 24.29 & 51.67 & 40.00 & 46.28 & 34.22 \\
UniAd-Base \cite{uniad} & CVPR 2023 & Ego State + 6 Cameras & 14.10 & 17.78 & 21.67 & 10.00 & 14.21 & 15.55 \\
ThinkTwice* \cite{thinktwice} & CVPR 2023 & Ego State + 6 Cameras & 27.38 & 18.42 & 35.82 & \textbf{50.00} & 54.23 & 37.17 \\
VAD \cite{vad} & ICCV 2023 & Ego State + 6 Cameras & 8.11 & 24.44 & 18.64 & 20.00 & 19.15 & 18.07 \\
DriveAdapter* \cite{driveadapter} & ICCV 2023 & Ego State + 6 Cameras & \textbf{28.82} & 26.38 & 48.76 & \textbf{50.00} & \textbf{56.43} & \textbf{42.08} \\
DriveTrans \cite{drivetransformer} & ICLR 2025 & Ego State + 6 Cameras & 17.57 & 35.00 & 48.36 & 40.00 & 52.10 & 38.60 \\
SparseDrive \cite{sparsedrive} & ICRA 2025 & Ego State + 6 Cameras & 12.50 & 17.50 & 20.00 & 20.00 & 23.03 & 18.60 \\
TTOG \cite{TTOG} & ArXiv 2025 & Ego State + 6 Cameras & 16.18 & 24.29 & 20.00 & 21.50 & 23.03 & 21.12 \\
\midrule 
GEMINUS* & {--} & Ego State + Front Camera & 11.11 & \textbf{37.50} & \textbf{55.00} & 40.00 & 45.26 & 37.77 \\
\bottomrule
\end{tabular}

\vspace{1ex} 
\begin{minipage}{\textwidth} 
\centering
\small
* denotes expert feature distillation.
\end{minipage}

\end{table*}

\begin{table}[t] 
\footnotesize 
\centering 

\setlength{\tabcolsep}{1pt} 

\caption{Ablation Study on Bench2Drive Benchmark.}
\label{tab3_ablation_study} 
\begin{tabular}{l|S[table-format=2.2]|S[table-format=2.2]|S[table-format=2.2]}
\toprule
\textbf{Method} & {DrivingScore} & {SuccessRate} & {MultiAbility} \\
\midrule
GEMINUS(\circled{1}+\circled{2}+\circled{3})                                  & \textbf{65.39} & \textbf{37.73} & \textbf{37.77} \\
ScenarioMoE-E2E (\circled{1}+\circled{2})             & 62.38          & 32.27          & 34.46          \\
VanillaMoE-E2E  (\circled{1}) & 59.23          & 29.09          & 32.05          \\
SingleExpert-E2E                              & 60.73          & 30.91          & 31.63          \\
\bottomrule
\end{tabular}

\vspace{1ex} 
\begin{minipage}{\columnwidth} 
\centering
\footnotesize 
\ For consistency, VanillaMoE-E2E uses five experts with Top-1 activation. \\
\circled{1} denotes Mixture-of-Experts. 
\circled{2} denotes scenario-aware routing. 
\circled{3} denotes uncertainty-aware routing and Global Expert.
\end{minipage}

\end{table}

\begin{table}[t]
\scriptsize
\centering

\setlength{\tabcolsep}{2pt}

\caption{Router Accuracy in Different Scenarios.}
\label{tab4 Router Accuracy}

\begin{tabular}{l|S[table-format=2.2]|S[table-format=2.2]|S[table-format=2.2]|S[table-format=2.2]|S[table-format=2.2]|S[table-format=2.2]}
\toprule
\textbf{Scenario} &  \textbf{{Overall}}  &  {Merging}  & {Overtaking} & {Em-Brake} & {Give Way} & {Traffic Sign} \\
\midrule
\textbf{Accuracy} & {\textbf{68.06\%}} & {32.85\%} & {91.35\%} & {54.03\%} & {2.87\%} & {90.45\%} \\
\bottomrule
\end{tabular}
\end{table}

\begin{table*}[t] 
\centering

\caption{Expert Utilization in Different Scenarios.} 
\label{tab5 Expert Utilization}

\setlength{\tabcolsep}{6pt}

\begin{tabular}{l|S|S|S|S|S|S}
\toprule
\textbf{Expert} & \multicolumn{6}{c}{\textbf{Expert Utilization (\%)}} \\
\cmidrule(lr){2-7}
\textbf{} & {Overall} & {Merging} & {Overtaking} & {Emergency Brake} & {Give Way} & {Traffic Sign} \\
\midrule
Global Expert & 6.29 & 6.43 & 1.09 & 10.52 & 6.70 & 6.04 \\
Merging Expert & 8.44 & \textbf{32.37} & 0.16 & 3.51 & 4.07 & 2.62 \\
Overtaking Expert & 19.22 & 3.13 & \textbf{91.04} & 1.75 & 61.72 & 0.84 \\
Em-Brake Expert & 16.08 & 3.67 & 5.66 & \textbf{52.33} & 15.07 & 5.03 \\
Give Way Expert & 0.23 & 0.00 & 0.31 & 0.24 & \textbf{2.15} & 0.17 \\
Traffic Sign Expert & 49.73 & 54.40 & 1.74 & 31.65 & 10.29 & \textbf{85.30} \\
\bottomrule
\end{tabular}
\end{table*}

\section{EXPERIMENTS}

\subsection{Experimental Setup}

{\textbf{Dataset.} GEMINUS is trained on the official Bench2Drive training dataset. This dataset is collected by Think2Drive \cite{think2drive}, a reinforcement learning expert with latent world model. To ensure a fair comparison with existing baselines, this paper utilizes the base dataset (1000 clips) for training and open-loop validation. This dataset comprises a 950-clip training set and a 50-clip open-loop validation set. Each clip represents a specific traffic scenario, spanning approximately 150 meters.

{\textbf{Evaluation Metrics.} For closed-loop evaluation, GEMINUS is assessed on 220 routes officially provided by Bench2Drive \cite{bench2drive}. These short routes are structured into 44 interactive scenarios, with 5 distinct routes per scenario. The closed-loop evaluation metrics include a Driving Score, Success Rate, and five MultiAbility metrics defined by Bench2Drive: Merging, Overtaking, Emergency Brake, Give Way, and Traffic Sign.

\textbf{Implementation Details.}
The model is trained for 32 epochs on a single NVIDIA GeForce RTX 4090 GPU with a batch size of 96. We use the Adam optimizer with an initial learning rate of $1\times10^{-4}$, a weight decay of $1\times10^{-7}$, and halve the learning rate after 30 epochs. The model takes $900 \times 256$ pixel RGB images to predict $K=4$ future waypoints at 2~Hz. Inspired by Bench2Drive, the training dataset is divided into five subsets (Merging, Overtaking, Emergency Brake, Give Way, and Traffic Sign) with corresponding scenario IDs $[0, 1, 2, 3, 4]$. The uncertainty threshold is set to $\tau=0.5$. We adopt PID parameters from Transfuser\cite{transfuser}, setting longitudinal ($K_P, K_I, K_D$) to (5.0, 0.5, 1.0) and lateral to (0.75, 0.75, 0.3). The loss weights are as follows: $\lambda_{traj}=1$, $\lambda_F=0.05$, $\lambda_V=0.001$, $\lambda_{Global}=1$, $\lambda_{Adaptive}=1$, $\lambda_{scenario}=1$, and $\lambda_{speed}=0.05$.

\subsection{Comparison with State-of-the-Art works}

As shown in Table \ref{tab1:Close and Open loop results}, GEMINUS achieves state-of-the-art performance on Bench2Drive closed-loop benchmark for both Driving Score and Success Rate. Notably, GEMINUS relies solely on monocular visual input, and surpasses existing methods on the Bench2Drive benchmark that use 6 camera.

While GEMINUS does not exhibit superior performance in open-loop average L2 error, such metrics primarily indicate model convergence rather than reliably assess real-world driving. In contrast, closed-loop metrics offer a more robust evaluation of actual driving capabilities, a point emphasized by previous research such as Transfuser++ \cite{hiddenbiaes} and Bench2Drive \cite{bench2drive}.

When focusing solely on monocular vision methods, GEMINUS significantly improves upon existing state-of-the-art monocular vision method, TCP-traj*\cite{tcp}. GEMINUS achieves a 9.17\% increase in Driving Score, a 25.77\% increase in Success Rate, and a reduction of 5.88\% in open-loop average L2 error. Furthermore, as shown in Table \ref{tab2 Multiability Results}, a 10.37\% increase in MultiAbility-Mean.

\subsection{Ablation Study}

As shown in Table \ref{tab3_ablation_study}, ablation study yields critical insights into the contribution of each GEMINUS component.

{\textbf{Comparing VanillaMoE-E2E with SingleExpert-E2E.} It is evident that directly introducing a generic MoE framework which is commonly used in LLMs into autonomous driving does not improve model performance. Without specific adaptation, it even leads to a slight decrease in Driving Score and Success Rate. This substantiates our hypothesis: end-to-end autonomous driving systems demand a more tailored MoE framework. Such a framework should specifically address the diverse and complex nature of real-world driving scenarios.

{\textbf{Comparing  ScenarioMoE-E2E with SingleExpert-E2E.} The scenario-aware routing mechanism comprehensively improves model performance. The Driving Score improved by 2.72\%, Success Rate by 4.40\%, and MultiAbility-Mean by 8.95\%. The introduction of this mechanism not only enhances the model's adaptive performance in diverse scenarios but also makes its routing logic more interpretable.

{\textbf{ Comparing GEMINUS with ScenarioMoE-E2E.} Further incorporating the uncertainty-aware routing mechanism and the Global Expert yields additional performance gains. The Driving Score improved by 4.83\%, Success Rate by 16.92\%, and MultiAbility-Mean by 9.61\%. The integration of the uncertainty-aware routing mechanism and the Global Expert significantly enhances the model's robustness and stability. This is particularly true in ambiguous scenarios where the router cannot confidently determine the current situation.

\begin{figure}[t]
\centering
\includegraphics[width=1\linewidth]{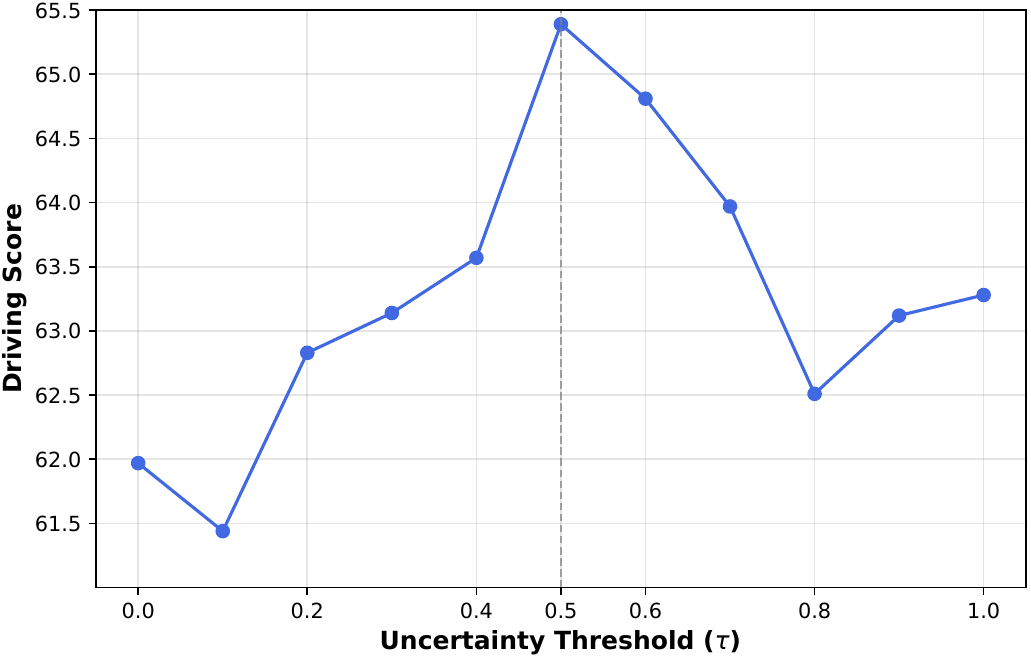}
\caption{\textbf{Driving Score variation trend with uncertainty threshold.}}
\label{DrivingScore trend}
\end{figure}

\subsection{Analysis of Uncertainty Threshold}

To investigate the impact of the uncertainty threshold $\tau$ on model performance, the uncertainty threshold $\tau$ is varied from 0.0 to 1.0 with a step size of 0.1 to conduct a series of closed-loop evaluations on the Bench2Drive Benchmark. As depicted in Fig.~\ref{DrivingScore trend}, the Driving Score shows a trend of initial increase followed by a decrease as $\tau$ gradually increases, reaching its optimum at $\tau$ = 0.5. This indicates that when the router's uncertainty is less than 0.5, the selection made by the scenario-aware routing is reliable, and the performance of the adaptive experts contributes to improved model performance. Conversely, when the router's uncertainty is greater than or equal to 0.5, the scenario-aware routing cannot make a reliable decision, necessitating the intervention of the Global Expert to ensure robust and stable performance.

\begin{figure}[t]
\centering
\includegraphics[width=1\linewidth]{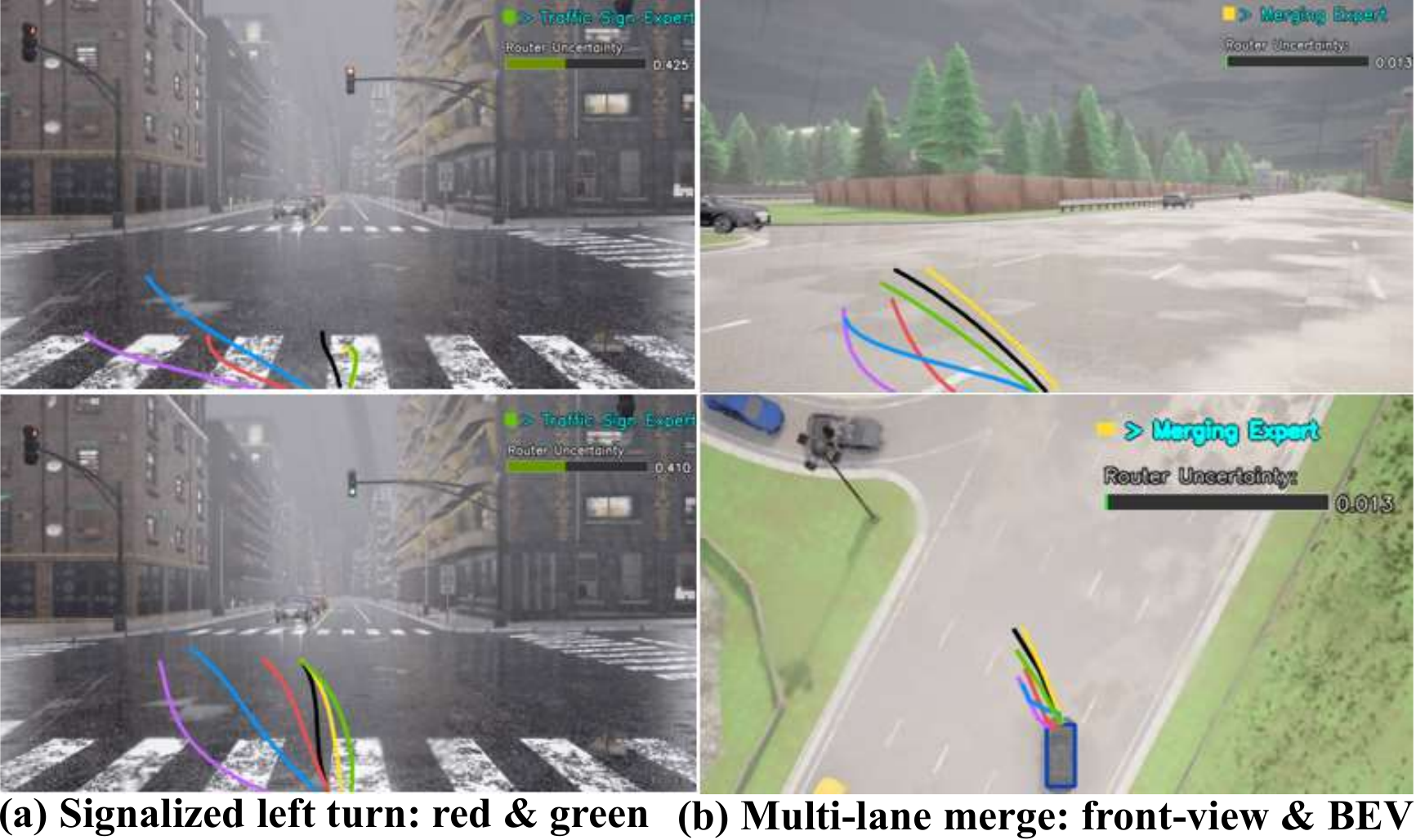}
\caption{\textbf{Qualitative examples of GEMINUS on the Bench2Drive closed-loop evaluation set.} 
For comparative purposes, while GEMINUS performs closed-loop control, we also visualize the predicted open-loop trajectories from other experts. These include the \textcolor{expertBlack}{Global}, \textcolor{expertYellow}{Merging}, \textcolor{expertMagenta}{Overtaking}, \textcolor{expertRed}{Emergency Brake}, \textcolor{expertBlue}{Give Way}, and \textcolor{expertGreen}{Traffic Sign} Expert.}
\label{Qualitative}
\end{figure}

\subsection{Router Accuracy and Expert Utilization}

To better understand the intrinsic routing dynamics of the GEMINUS framework, analysis is conducted on the Bench2Drive open-loop validation set. This analysis focused on two key aspects during open-loop evaluation: router prediction accuracy and expert utilization. Router prediction accuracy is defined as the proportion of samples where the router correctly identifies the corresponding scenario. Expert utilization refers to the activation rates of both the Global Expert and the five Scene-Adaptive Experts.

{\textbf{Router Accuracy.} As depicted in Table~\ref{tab4 Router Accuracy}, the router's overall scenario prediction accuracy reached 68.06\%. It is worth noting that the Traffic Sign subset overlaps with both the Merging and Emergency Brake subsets. In such cases, a single sample pertains to two scenario types. Therefore, the actual prediction accuracy could be even higher. This indicates that the scenario-aware routing  can accurately determine the current scenario in most cases. However, it struggles with an accurate prediction in a minority of scenarios. A closer examination of the five validation set scenarios reveals that, in the Overtaking and Traffic Sign scenarios, the router exhibits the highest prediction accuracy. This is mainly because these scenarios have salient visual cues, such as obstacles or traffic signs. These cues significantly enhance the router's ability to accurately predict the scenario. In contrast, the Give Way scenario presents the lowest prediction accuracy of 2.87\%. This discrepancy stems from two primary factors. First, the Give Way subset constitutes only 3.2\% of Bench2Drive-base dataset. This represents an inherent data imbalance within the official Bench2Drive dataset. Second, GEMINUS relies on monocular visual input. This constrains its ability to detect rear-approaching vehicles in Give Way scenarios, thereby impeding accurate scenario prediction.

{\textbf{Expert Utilization.} As depicted in Table \ref{tab5 Expert Utilization}, the “Overall” column reveals a Global Expert utilization rate of 6.29\%. This indicates that GEMINUS primarily prioritizes routing to scene-adaptive experts in most instances. This allows it to leverage their scenario-specific capabilities. The Global Expert is mainly invoked only in highly ambiguous scenarios to ensure robust and stable performance. Furthermore, a comparative analysis of the “Global Expert” row in Table \ref{tab5 Expert Utilization} with the router accuracy in Table \ref{tab4 Router Accuracy} shows a clear pattern. Global Expert utilization is minimal in scenarios with higher routing prediction accuracy, such as Overtaking (1.09\%) and Traffic Sign (6.04\%). Conversely, in the three scenarios characterized by lower routing prediction accuracy, the model exhibits increased Global Expert utilization. This helps to maintain robustness and stable performance.

\subsection{Qualitative Results}

Fig.~\ref{Qualitative} presents qualitative results for GEMINUS in two complex closed-loop scenarios, showcasing the effectiveness of our Dual-aware Router.

In the signalized left turn scenario (a), when facing a red light, the router correctly selects the Traffic Sign Expert. This choice is critical for safety, as the visualization shows that several other experts predict dangerous trajectories that would proceed illegally through the intersection. In stark contrast, the chosen expert’s trajectory (green) demonstrates correct compliance by stopping safely before the crosswalk. Subsequently, when the light turns green, it executes a smooth and safe turning arc.

Similarly, in the multi-lane merge (b), GEMINUS activates the Merging Expert. The resulting trajectory (yellow) is distinguished by a well-calibrated curvature and the correct directional heading, making it the ideal path for a smooth and effective merge.

\section{CONCLUSIONS}

This paper presents GEMINUS, a novel Dual-aware Mixture-of-Experts framework tailored for end-to-end autonomous driving. Through the effective coupling of the Global Expert and Scene-Adaptive Experts Group via the Dual-aware router, GEMINUS simultaneously achieves adaptive and robust performance in complex and diverse scenarios. Comprehensive evaluations on the Bench2Drive benchmark substantiate the framework's effectiveness, where GEMINUS achieves state-of-the-art performance and demonstrates significant improvements over baseline methods, relying solely on monocular visual input.

This study is limited by the use of monocular camera inputs. To enable the router to consider scene information more comprehensively, the exploration of Dual-aware Router with multi-camera input remains a promising direction for future research. Additionally, our adherence to the predefined Bench2Drive scenario categories introduces challenges like data imbalance and scenario overlaps, which impede expert specialization and thus limit the full potential of GEMINUS. Future work could focus on designing more rational and fine-grained scenario classification criteria. Furthermore, a promising research direction is to replace GEMINUS's expert networks with Low-rank Adaptation (LoRA) modules, providing a lightweight Mixture-of-Experts plugin that is highly effective for the efficient fine-tuning of pretrained models.






\bibliographystyle{IEEEtran}

\end{document}